%% file: emnlp2024.tex
\pdfoutput=1

\documentclass[11pt]{article}

\usepackage[final]{acl}

\usepackage{times}
\usepackage{latexsym}
\usepackage{algorithm}
\usepackage{mathrsfs}
\usepackage{enumitem}
\usepackage{algpseudocode}

\usepackage[T1]{fontenc}
\usepackage{url}            
\usepackage{booktabs}       
\usepackage{amsfonts}       
\usepackage{nicefrac}       
\usepackage{microtype}      
\usepackage{xcolor}         
\usepackage{amsmath}  
\usepackage{graphicx}  
\usepackage{subcaption} 
\usepackage{multirow}
\usepackage{wrapfig}
\usepackage{array}
\usepackage{bm}
\usepackage[utf8]{inputenc}
\usepackage{xspace}

\newcommand{\our}{\text{PromptIntern}\xspace}
\usepackage{microtype}

\usepackage{inconsolata}

\usepackage{colortbl} 
\usepackage{graphicx}

\title{\our: Saving Inference Costs by Internalizing Recurrent \\Prompt during Large Language Model Fine-tuning

}
\makeatletter
    \def\@fnsymbol#1{\ensuremath{\ifcase#1\or \dagger\or \ddagger\or
       \mathsection\or \mathparagraph\or \|\or **\or \dagger\dagger
       \or \ddagger\ddagger \else\@ctrerr\fi}}
\makeatother
    
\author{
Jiaru Zou\textsuperscript{\rm 1}\thanks{\indent The contributions by Jiaru Zou and Tao Li have been conducted and completed during their internships at Microsoft.}, 
Mengyu Zhou\textsuperscript{\rm 2}\thanks{\indent Corresponding author.}, 
Tao Li\textsuperscript{\rm 3}\footnotemark[1], 
Shi Han\textsuperscript{\rm 2}, 
Dongmei Zhang\textsuperscript{\rm 2} \\
\textsuperscript{\rm 1} University of Illinois Urbana-Champaign
\textsuperscript{\rm 2} Microsoft \\
\textsuperscript{\rm 3} Shanghai Jiao Tong University \\
\texttt{\href{mailto:jiaruz2@illinois.edu}{jiaruz2@illinois.edu}},
\texttt{\href{mailto:li.tao@sjtu.edu.cn}{li.tao@sjtu.edu.cn}}, \\
\texttt{\{\href{mailto:mezho@microsoft.com}{mezho}, \href{mailto:shihan@microsoft.com}{shihan}, \href{mailto:dongmeiz@microsoft.com}{dongmeiz}\}@microsoft.com}
}

\begin{document}
\maketitle
\begin{abstract}
Recent advances in fine-tuning large language models (LLMs) have greatly enhanced their usage in domain-specific tasks. Despite the success, fine-tuning continues to rely on repeated and lengthy prompts, which escalate computational expenses, require more resources, and lead to slower inference. In this paper, we present a novel approach, \our, which internalizes prompt knowledge during model fine-tuning to achieve efficient inference and save costs. Instead of compressing the prompts for a vanilla model, \our aims to embed the recurrent prompt directly into the model parameters. We design a fine-tuning pipeline that includes instruction template compression, few-shot example absorption, and a progressive internalization strategy, effectively diminishing the need for intricate prompts during inference. Comprehensive experiments on challenging NL2Code tasks demonstrate that our method reduces input tokens by more than 90\%, accelerates inference by 4.2 times, and reduces monetary inference costs by 88.3\%.

\end{abstract}

\input{source/1_introduction}

\input{source/2_relatedwork}
\input{source/3_method}
\input{source/4_experiment}

\input{source/6_discussion}
\input{source/5_conclusion}

\newpage
\bibliography{emnlp2024}

\appendix
\input{source/appendix}

\end{document}

%% file: source/1_introduction.tex
\section{Introduction}
Large language models (LLMs) have become pivotal in numerous natural language processing (NLP) applications, such as natural language generation~\cite{dong2019unified, zheng2024heterogeneous}, reasoning ~\cite{zhu2023minigpt,sui2023tap4llm}, and code generation~\cite{luo2023wizardcoder,he2024conline, rozière2024code}. 
To enhance the predictive accuracy of LLMs in domain-specific tasks, recent techniques in fine-tuning, such as parameter-efficient fine-tuning (PEFT) \cite{he2021towards, hu2021lora, lester2021power}, have been developed for pre-trained models to excel in specific tasks by adjusting their parameters to better align with targeted datasets \cite{hu2021lora}. Many of these fine-tuning approaches typically adopt prompts that are optimized and integrated with detailed instructions, examples, and retrieved documents through prompt engineering techniques such as chain-of-thought \cite{wei2022chain}, few-shot prompting \cite{brown2020language}, and retrieval-augmented generation \cite{lewis2020retrieval, cheng2023batch}.

\begin{figure}[!t]
    \centering
    \includegraphics[width=0.9\linewidth]{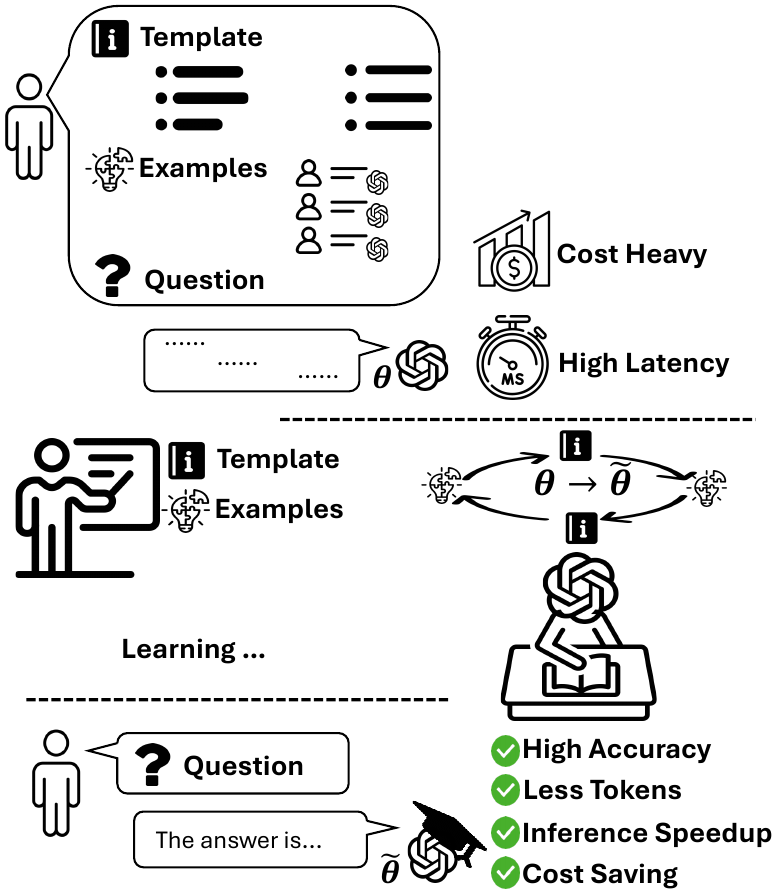}
    \caption{{An illustration of \our: Like human interns, LLMs learn and internalize repeated prompt information such as templates and examples during fine-tuning, leading to efficient and effective inference.}}
    \label{fig:illustration-1}
    \vspace{-3mm}
\end{figure}

Although these advancements enhance the capabilities of LLMs during fine-tuning, they also present new challenges: Prompt engineering often necessitates longer prompts, and directly integrating lengthy prompts into the training process further increases computational costs during inference \cite{vm2024fine}. 
This increase in cost precludes LLMs in many cost-sensitive scenarios where computational resources are constrained. 
Several prompt compression methods \cite{li2023compressing,jiang2023llmlingua,pan2024llmlingua} have been proposed to reduce text redundancy. 
They design various prompt compression systems and strive to preserve maximum information between original and compressed prompts. While these methods ensure the retention of original prompt information, they mainly focus on the prompt perplexity and overlook the adaption of target LLMs during compression \cite{pan2024llmlingua}. 
For challenging tasks that require model fine-tuning \cite{mosbach2023few}, these approaches struggle to establish connections between compressed tokens and dynamically adjusted model parameters. Consequently, naively applying these compression methods often leads to large performance degradation, as relevant information may be inadvertently removed or distorted during the compression process.

In this paper, we propose a novel prompt internalization approach, namely \textbf{\our}, which internalizes prompt input during model fine-tuning and enables efficient inference. Unlike prompt compression which removes tokens based solely on prompt information entropy, we aim to transfer various types of prompt knowledge into updated model parameters, thereby directly enhancing LLMs' understanding. Our idea is motivated by the human learning process as illustrated in Figure~\ref{fig:illustration-1}: During internship on-boarding, human interns need detailed instructions, examples, and documents to learn new tasks. As they internalize these materials and become familiar with their duties, they master the necessary skills and no longer require extra guidance.
Similarly, when specific prompt information (e.g. task constraints, output formats/styles) is repeatedly exposed to an LLM during fine-tuning, the model can gradually internalize the knowledge into its updated parameters. Such repeated information can be progressively eliminated from prompt inputs since it becomes unnecessary for inference of the master LLM.

We dub our approach \textbf{\our} to regard LLMs as human \textbf{intern}s and \textbf{intern}alize prompt knowledge progressively. 
Our approach consists of several key steps: Initially, we classify an input prompt into three components: the template, examples, and query. We start by setting a schedule to linearly decrease both the template compression rate and the number of few-shot examples across training stages. Following the schedule, we implement template compression and example absorption to pre-process the input prompts. We then introduce a comprehensive pipeline that enables LLMs to progressively internalize template and example components into model parameters during fine-tuning and efficiently perform inference using query-only prompts.

We assess our method on challenging NL2Code tasks \cite{zan2022large} that are widely recognized as benchmarks for model fine-tuning. 
Our experiments evaluate \our on three key metrics: accuracy, token usage, and inference speed. The results indicate that under identical fine-tuning settings, our method not only surpasses prompt compression methods but also achieves comparable accuracy to direct fine-tuning. Moreover, it accelerates the inference process by a factor of 4.2 and reduces token usage by over 90\% compared to direct fine-tuning. These enhancements demonstrate that our approach successfully balances efficiency and effectiveness, making it well-suited for optimizing LLM performance across various cost-saving scenarios. We further quantify the total monetary cost savings and conduct detailed analyses to elucidate the efficacy of our approach and provide insights into its underlying mechanisms. 
Our main contributions can be summarized as follows:
\begin{itemize}[leftmargin=*,itemsep=0pt]
    \item We proposed \emph{\our}\footnote{Our code will be released at \url{https://github.com/microsoft/PromptIntern}}, a novel prompt internalization method that aims to internalize repetitive prompt knowledge into the model's parameters, achieving high inference efficiency while maintaining model performance.
    \item We devised detailed prompt internalization strategies for template compression and example absorption along with a tailored progressive fine-tuning pipeline.
    \item We conducted extensive experiments with detailed analyses on challenging NL2Code tasks. The experimental results show that our approach reduces token usage by over 90\%, speeds up inference time by 4.2 times, and achieves 88.3\% cost savings across a broad spectrum of LLMs.
\end{itemize}

%% file: source/2_relatedwork.tex
\section{Related Work}
\begin{figure*}[!t]
    \centering
    \includegraphics[width=1\linewidth]{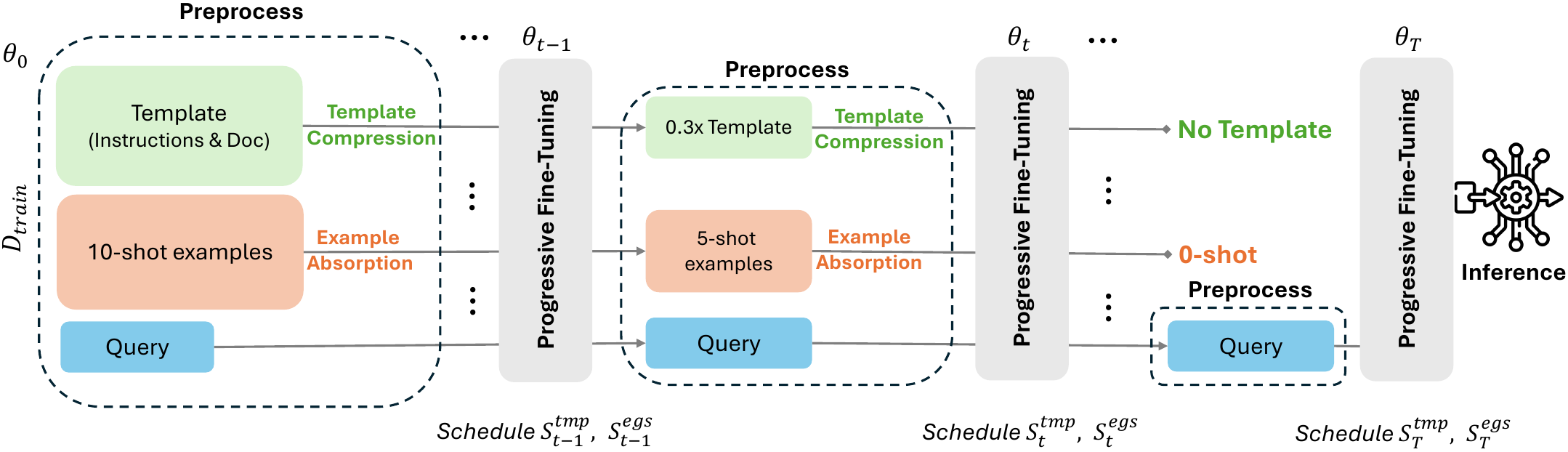}
    \caption{{Overview of \our framework. We structure the input prompt into three components: the template, examples, and query. By employing template compression and example absorption, we efficiently preprocess each component based on schedule $\mathcal{S}^{tmp},\mathcal{S}^{egs}$. We then use a progressive fine-tuning strategy to gradually incorporate prompt knowledge into the model parameters $\theta$, facilitating efficient inference without sacrificing performance.
    }}
    \label{fig:framework}
\end{figure*}
\textbf{Prompt compression} rephrases original prompts more concisely and is classified into task-aware and task-agnostic approaches. Specifically, the task-aware approaches, like LongLLMLingua \cite{jiang2023longllmlingua}, utilize a question-aware coarse-to-fine-grained strategy to compress information based on the query. In contrast, methods like soft prompts \cite{wingate2022prompt, liu2022p, mu2024learning} use learnable tokens to condense prompts. Conversely, the task-agnostic methods utilize metrics such as information entropy to eliminate redundant prompt information, with systems like LLMLingua \cite{jiang2023llmlingua, li2023compressing} estimating token importance using a smaller model. Despite the demonstrated effectiveness of these methods, producing compressed text that can generalize across different tasks and be effectively integrated into training scenarios remains a challenge.
\\ \\
\noindent\textbf{Model fine-tuning} adopts pre-trained LLMs to specific tasks by modifying parameters. Based on the assumption that fine-tuning adds less new information to the model pre-trained on large internet-scale datasets, Parameter-Efficient Fine-Tuning (PEFT) methods aim to curtail the costs of tuning large models by adjusting a subset of parameters. Existing PEFT methods can be broadly categorized into three main approaches: 1) Adapter-based methods \cite{houlsby2019parameter,he2021towards}: Introduce trainable modules within a static "backbone" network, offering flexibility but potentially increasing model size. 2) Prompt-based methods \cite{lesterpower,razdaibiedina2023residual,nashid2023retrieval}: Employ trainable "soft tokens" at input sequence start, requiring effective prompt design per task. 3) Low-rank adaptation methods \cite{hu2021lora,dettmers2024qlora,liu2024dora}: Use low-rank matrices to approximate required weight adjustments, avoiding additional inference burden and often delivering strong performance. Despite advancements in fine-tuning strategies, data inputs should be carefully managed and distinguished from lengthy ones used for direct inference.

%% file: source/3_method.tex
\section{Problem Formulation} 
Let an input prompt as \( x = (x^{tmp}, x^{egs}, x^{que}) \), where each input prompt \( x \) is considered as a tuple of three components: $x^{tmp}$ as the template such as fixed instructions, API docs, etc., $x^{egs}$ as the examples, and $x^{que}$ as the query. Typically, $x^{tmp}$ and $x^{egs}$ are relatively fixed and lengthy but essential for complex tasks.
Let \( f_\theta(\cdot) \) denote the neural network function of a LLM model, typically transformer~\cite{vaswani2017attention}, parameterized by \( \theta \). The generated output by LLM can be represented as $f_\theta(x)$. 

We then consider the following problem of prompt internalization. Given a training dataset \( \mathcal{D}_{train}=\{ (x_i,y_i) \}_{i=1}^n \) where $n$ is the number of training samples, \(x_i\) is an input prompt defined above, and $y_i$ is the corresponding groundtruth output.
Our goal is to internalize the knowledge contained in templates and examples of each input prompt i.e. \( \{ (x_i^{tmp}, x_i^{egs}) \}_{i=1}^n \) into model parameters $\theta$ during fine-tuning, enabling efficient inference while maintaining high prediction performance through \( \{x_i^{que}\}_{i=1}^n \) only.
Formally, the prompt internalization objective can be formulated as follows:
\begin{equation}
    \min_{\tilde{\theta}} \sum_{i=1}^n \mathcal{L}\left ( y_i, f_{\tilde{\theta}}( x^{que}_i)\right)
\end{equation}
where $\mathcal{L}(\cdot)$ denotes the loss function and $\tilde{\theta}$ denotes the updated weights with internalized prompt knowledge. For a new incoming prompt only containing the query, the updated LLM with $f_{\tilde{\theta}}( \cdot)$ can internally recover the output without the assistance of instruction and examples.

\section{Methodology}
In this section, we introduce our method \our in detail. We first present the template compression to compress the entire fixed template part inside a prompt. Then we show the example absorption to effectively absorb demonstration examples into model parameters. Finally, we introduce a tailored training strategy for \our. The overall framework is shown in Figure \ref{fig:framework}.


\subsection{Template Compression}
We first introduce template compression, which is designed to compress the common template information exists across training instances. The motivation of the template compression stems from the following aspects: 1) Redundancy. The instruction is repetitive across prompts for a given task, often containing unnecessary tokens that do not contribute to the language model's understanding, posing significant memory and computational burdens when the instruction is lengthy; and 2) Noise. Excessively long prompts may incorporate extraneous elements—either irrelevant or misleading information—that serve as noise and can adversely affect the model's generation.


To mitigate the issues stated above, we propose a template compression system, which can generally be expressed as:
\begin{equation} \label{eqa:template_compression}
   \tilde{x}^{\text{tmp}} = C(x^{\text{tmp}}, \tau^{\text{tmp}})
\end{equation}
where \(C\) is a specific template compressor, \(\tilde{x}^{tmp}\) is the compressed template, and \(\tau^{tmp}\) is the template compression rate as defined in \cite{jiang2023llmlingua}, varying at differnt training interations. 
We then adopt a predetermined schedule $\mathcal{S}^{tmp}(t)$ to progressively reduce and internalize the prompt template information during the $t$-th training iteration. Specifically, for a total of $T$ training iterations, we initially set $\tau^{tmp}$ to 1 at $\mathcal{S}^{tmp}(0)$ and gradually decrease the value of $\tau^{tmp}$ at $\mathcal{S}^{tmp}(t)$ to zero at end to achieve fully template internalization. Note that such a compression system is also flexible, allowing it to halt at a desired non-zero compression rate. This flexibility allows to maintenance of a certain level of compressed template, serving as a trade-off to preserve inference accuracy in specific scenarios, as discussed in Section \ref{sec:ablation}. In addition to the progressively decreasing template schedule, we also specify the template compressor $C$ for better utilization. we categorize it into two types which exactly reflect the primary components of the template defined in the problem formulation: the instruction compressor and document compressor: 

\textit{Instruction Compressor} targets the static elements within prompts, specifically focusing on the instructional content. Instructions in training data often consist of repeated directives, guidelines, or predefined tasks which are common across multiple training scenarios. The primary goal of the instruction compressor is to distill these instructions down to their essential components, eliminating verbosity and redundancy without compromising the clarity or intent of the instructions. 

\textit{Document Compressor} is designed to handle the bulkier and more detailed portions of the prompts, such as API documentation or static demonstrations. These sections typically include extensive technical descriptions and examples that, while informative, often contain a significant amount of repetitive or non-essential information \cite{xu2023recomp}. The goal of the document compressor is to reduce the information unnecessary for understanding and applying the technical content, thereby streamlining the training process. 
\subsection{Example Absorption}
Incorporating few-shot examples into fine-tuning not only improves information retrieval and memory recall~\cite{hubotter2024active} but also yields substantial benefits in handling a variety of tasks with minimal data input~\cite{mosbach2023few, snell2017prototypical}.
However, directly adding lengthy few-shot examples to input prompts burdens the context window and increases inference latency. Motivated by this, we propose example absorption to benefit from the enhanced performance afforded by few-shot examples while preventing incurring significant additional overhead.
Specifically, the example absorption mainly contains two stages: example retrieval and example removal. 

\textit{Example Retrieval} is designed to identify and select the most related few-shot examples from the training dataset and incorporate them into each training instance. The underlying rationale is to choose examples that closely align with the training instance so as to accelerate model's internalization during training. 
We employ a straightforward approach that utilizes a relevance scoring function $s(\cdot, \cdot)$ to assess the similarity between examples and the training instance. Specifically, we select the top $k$ examples, varying at different training iterations, with the highest relevance scores to serve as our few-shot examples. For a training instance $(x_i, y_i)$ with $x_i$ being the input prompt and $y_i$ being the corresponding groundtruth output, the selection process can be expressed as follows:
\begin{align} \label{eqa:example_retrieval}
x^{egs}_i = \left\{ (x_j, y_j) \mid j \neq i, s(y_i, y_j) \in \text{top } k\text{ scores} \right\}
\end{align}
Note that the scoring function is calculated based on common similarity metrics ~\cite{rubin2022learning,chen2022decoupling,dai2022promptagator}. In our experiment, we use the BLEU as the scoring function.


\textit{Example Removal} aims to progressively internalize the prompt knowledge from few-shot examples into model parameters. To achieve this, we also adopt a predetermined schedule $\mathcal{S}^{egs}(t)$ to gradually decrease the number of demonstration examples in each prompt instance during the t-th iteration. Specifically, for a total of $T$ training iterations, we initially set $k$ examples at $\mathcal{S}^{egs}(0)$ and then gradually decrease the value of $k$ at each $\mathcal{S}^{egs}(t)$ to zero at end in order to achieve fully example internalization.


\subsection{\our Pipeline}

\noindent
In this subsection, we describe the detailed pipeline of \our. As demonstrated in Algorithm \ref{alg:inprompt_training},  \our consists of three stages: preprocess (line 1-7), progressive fine-tuning (line 8-12), and inference (line 13-14).

\textit{Preprocess.} For the first step, We preprocess the input prompts to prepare them for the progressive training stage. Specifically, we process the prompt template to different compression rates based on the schedule $\mathcal{S}^{tmp}(t)$ and retrieve examples for each training instance based on the schedule $\mathcal{S}^{egs}(t)$. For better illustration, we provide an example of a pre-processed prompt with respect to schedule in Appendix \ref{appendix: example_demo}.
\begin{algorithm}[!t]
\renewcommand{\algorithmicrequire}{\textbf{Input:}}
\renewcommand{\algorithmicensure}{\textbf{Output:}}
\caption{\our Pipeline}\label{alg:inprompt_training}
\begin{algorithmic}[1] 
\Require  A training dataset 
$\mathcal{D}_{train}\!=\!\!\{(x_i, y_i)\}_{i=1}^n$ 
with $x_i=(x^{tmp}_i, x^{egs}_i, x^{que}_i)$ and corresponding labels $y_i$, A language model $f$ with initial parameters $\theta$, learning rate $\eta$, training iterations $T$, template compression schedule $\mathcal{S}^{tmp}$, example absorption schedule $\mathcal{S}^{egs}$
\Ensure The inference output $f_{\theta_T} (x^{que})$
\State \underline{\textit{Preprocess}}
\For{ $i = 1, 2, \dots, n$} 
\State Obtain each $\tau^{tmp}$ from $\mathcal{S}^{tmp}$
\State Obtain each $k$ from $\mathcal{S}^{egs}$
\State Compress $x^{tmp}_i$ w/ each $\tau^{tmp}$ via Eq.~(\ref{eqa:template_compression})
\State Retrieve $k$ examples $x^{egs}_i$ via Eq.~(\ref{eqa:example_retrieval})
\EndFor
\State \underline{\textit{Progressive Finetuning}}
\For{ $t = 0, 1, \dots, T-1$} 
\State Adjust prompts with $\mathcal{S}^{tmp}(t)$ and $\mathcal{S}^{egs}(t)$
\State Update model parameters $\theta_t$ via Eq.~(\ref{eqa:fine-tuning})
\EndFor
\State \underline{\textit{Inference}}
\State Perform inference with $f_{\theta_T} (x^{que})$
\end{algorithmic}
\end{algorithm}

\input{source/tables/compression_based} 
\textit{Progressive Fine-tuning.} We then fine-tune the model parameters for internalizing. Given the training iteration $t$, we update the model parameters as follows:
\begin{align}
\begin{split}
\label{eqa:fine-tuning}
{\theta}_{t+1} = {\theta}_t - \frac{\eta}{b} \sum_{i=1}^b \nabla_\theta \mathcal{L} \big (f_{\theta_t}(x_i^{tmp}(t),   \\   x_i^{egs}(t), x_i^{que}), y_i \big )
\end{split}
\end{align}
where $\eta$ is the learning rate, $\mathcal{L}$ is the cross-entropy loss function, $b$ is the batch size, $\mathcal{B}=\left \{ (x_i, y_i) \right \}_{i=1}^b$ is the data batch, and $y$ is the groundtruth label.

\textit{Inference.} After the progressive fine-tuning, we have trained the LLMs with updated model parameters $\theta_T$ to perform inference without adding instructions or any examples. Thus, we can predict the output simply with $f_{\theta_T} (x^{que})$.

Our objective is to effectively compress and incorporate prompt knowledge into model parameters that are specifically tailored for distinct tasks. In pursuit of this goal, we have adopted PEFT during the fine-tuning phase of \our. Specifically, we apply LoRA ~\cite{hu2021lora} as it imposes no additional computational costs during inference and allows for scalable deployment across multiple tasks ~\cite{sheng2023s}. Note that our outlined pipeline in Algorithm~\ref{alg:inprompt_training} is also compatible with other PEFT techniques.

%% file: source/tables/compression_based.tex
\begin{table*}[!t]
  \centering
  \caption{Comparison with prompt compression baselines on NL2Code benchmark. To ensure a fair comparison, we apply each baseline to compress the input prompt and use the compressed prompt as both training and testing data during model fine-tuning. We also standardize the compression ratio across all methods to approximately the same as indicated by $1/\tau_{all}$.}
  \label{tab:baseline_compare}
  \resizebox{\textwidth}{!}{
    \begin{tabular}{lccccccccc}
    \toprule
    \textbf{Methods} & \multicolumn{3}{c}{MBPP} & \multicolumn{3}{c}{NL2F} & \multicolumn{3}{c}{NL2Bash} \\
    \cmidrule(lr){2-4} \cmidrule(lr){5-7} \cmidrule(lr){8-10}
    (\textit{Inference on GPT-3.5}) & Pass@1 & Input Tokens & $1/\tau_{all}$ & E.M. & Input Tokens & $1/\tau_{all}$ & BLEU & Input Tokens & $1/\tau_{all}$ \\
    \midrule
    GPT4 Generation & 61.8 & 128 & 1.8x & 59.6 & 425 & 1.6x & 59.5 & 256 & 1.9x \\
    Selective Context & 59.7 & 102 & 2.2x & 56.4 & 391 & 1.7x & 55.2 & 158 &  3.1x \\
    LLMLingua & 70.3 & 115 & 2.0x & 64.2 & 417 & 1.6x & 61.3 & 154 &  3.1x \\
    LongLLMLingua & 65.2 & 121 & 1.9x & 67.8 & 425 & 1.6x & 58.4 & 133 &  3.6x \\
    LLMLingua-2 & 72.5 & 107 & 2.1x & 70.4 & 407 & 1.7x & 62.8 & 141 &  3.4x \\
    \midrule
    \textbf{\our} & \textbf{78.1} & 107 & 2.1x & \textbf{81.4} & 407 & 1.7x & \textbf{70.5} & 141 & 3.4x\\
    \bottomrule
    \end{tabular}
  }
\end{table*}

    

%% file: source/4_experiment.tex
\section{Experiment}
In this section, we evaluate the performance of \our across various benchmarks on the NL2Code task. The NL2Code task is widely recognized for its utility in evaluating LLMs on both fine-tuning efficacy and cost-effectiveness in real-world applications \cite{zan2022large}. Following this, our experiments primarily focus on two key perspectives: \textbf{1) Effectiveness}: assessing the performance accuracy of \our during inference phases; \textbf{2) Efficiency}: quantifying the reduction in token usage and corresponding cost savings achievable through \our.   

\subsection{Settings}
\paragraph{Datasets} We apply three typical NL2Code datasets: MBPP~\cite{austin2021program} for NL to python code generalization, NL2F~\cite{zhao2024nl2formula} for NL to Excel spreadsheet formulas generation, NL2Bash~\cite{lin2018nl2bash} for NL to Bash Shell commands generation. Please refer to Appendix \ref{appendix:dataset} for the dataset details.

\paragraph{Evaluation Metrics}
We use one-shot pass accuracy $Pass@1$ \cite{austin2021program} for MBPP,  \textit{Exact Match (E.M.)} for NL2F, and \textit{BLEU} score for NL2Bash. We also calculate the input tokens usage and compression ratio $1/\tau$ for each dataset.

\paragraph{Baselines} We consider two types of baselines with setups below: 

\noindent
1) \textit{Prompt Compression approaches.} We employ the latest advancements in prompt compression techniques. Specifically, we utilize Gist Tokens~\cite{mu2024learning}, GPT-4 Generation~\cite{jiang2023longllmlingua}, Selective Context\cite{li2023compressing}, and LLMLingua series ~\cite{jiang2023llmlingua, jiang2023longllmlingua, pan2024llmlingua}. Each prompt compression method is initially applied to compress the entire dataset to a predetermined compression rate. Then, the compressed dataset is utilized for both fine-tuning and inference evaluation.

\noindent
2) \textit{Direct Fine-tuning approaches.} We use ``Direct'' as the counterpart to our progressive fine-tuning strategy. Specifically, we adopt several conventional direct fine-tuning configurations, including i) direct fine-tuning with complete template and examples (e.g. \emph{Template with 5-shots} in Table~\ref{tab:direct_ft res}), ii) direct fine-tuning with compressed template and reduced examples (e.g. \emph{Template x0.6 with 2-shots} in Table~\ref{tab:direct_ft res}), iii) direct fine-tuning with template only (\emph{Template only}), and iv) direct fine-tuning without template and examples (\emph{No template}).

\noindent
\paragraph{Models} To demonstrate the broad applicability of \our, we utilize both closed-source and open-source LLMs with different parameter sizes for fine-tuning and inference processes.1) Closed-Source: We apply GPT-4-0613 \cite{openai2023gpt}, abbreviated as GPT-4, and GPT-3.5-turbo-0125\footnote{https://platform.openai.com/docs/models/gpt-3-5-turbo}, abbreviated as GPT-3.5. 2) Open-Source: We apply Mixtral-8x7B-v0.1 \cite{jiang2024mixtral}, abbreviated as Mixtral-8x7B, Llama2-7B \cite{touvron2023llama}, and Llama2-13B \cite{touvron2023llama}.

\paragraph{Implementation Details} Please refer to Appendix \ref{appendix:add_exp} for the additional experiments settings and implementation details.
\input{source/tables/direct_ft_res}
\noindent
\subsection{Prompt Compression Comparison}
Table \ref{tab:baseline_compare} reports the overall result of \our with the prompt compression baselines inferenced on GPT-3.5 across all datasets. Here we establish the template compression rate $\tau_{tmp}$ at 0.3 across all prompt compression approaches as well as \our to ensure a fair comparison. And $\tau_{all}$ in the table represents the overall prompt's compression rate.  
We observe that while utilizing a comparable number of tokens for inference, our approach outperforms all baselines, achieving improvements of 5.6\% on MBPP, 11.0\% on NL2F, and 7.7\% on NL2Bash. The result demonstrates that \our generally offers the best balance of efficiency and effectiveness across varied tasks. Note that since the Gist Token\cite{mu2024learning} baseline is only applicable to open-source LLMs, we separately compare it with our approach which can be found in Appendix \ref{appendix:gist}.

\subsection{Direct Fine-tuning Comparison}
Table \ref{tab:direct_ft res} shows the comparison of our approach with direct fine-tuning baselines on three datasets. In MBPP, our method outperforms the \emph{No Template} and \emph{Template} baselines by 9.6\%-10.8\% and 0.6\%-1.3\%, respectively, and achieves similar results to \emph{Template x0.6 with 2-shots}, using fewer tokens. We also reduce input token usage by 9.8x-12.2x, compared to the \emph{Template with 5-shots}, which requires 22.2x-27.4x more tokens.
In NL2F, our method improves over \emph{No Template} by 8.0\%-19.0\% with the same token usage, matching the \emph{Template x0.6 with 2-shots} baseline while reducing token usage by 6.4x. We also observe that larger models (LLama2-13B) show less degradation without prompt templates compared to smaller models (LLama2-7B), with a difference of -5.7\% verses -21.2\%.
In NL2Bash, our approach exceeds \emph{No Template} by 7.3\%-11.3\% and reduces token usage by 15.5x compared to the baseline method \emph{Template x0.6 with 5-shots}.

\input{source/tables/ablation_study/variants}
\input{source/tables/ablation_study/schedule}
\subsection{Ablation Study} \label{sec:ablation}
To effectively assess the impact of various components within \textit{\our}, we introduce three variants of \our for ablation studies:
\begin{itemize}[leftmargin=*,itemsep=0pt] 
    \item \textbf{\our w/\(\tau_{tmp}\)=0.3}, where we set the compression rate to 0.3 instead of 0 in template compression.
    \item \textbf{\our w/o Example Absorption}, where we omit the example absorption for retrieving and internalizing few-shot examples during fine-tuning.
    \item \textbf{\our w/o Template Compression}, where template compression is excluded for both fine-tuning and inference prompt instances.
\end{itemize}
\noindent
The overall results is shown in Table \ref{tab:ablation_study_}. When comparing \our with \our w/$\tau_{tmp}=0.3$, we observe an average of 2.4\% drop on performance but a 3.7x compression ratio on token usages across all three datasets. This highlights the balance between compression rate and accuracy performance. When comparing \textit{our} with \textit{our} w/o Example Absorption, we observe a large performance drop in the latter variant, despite both approaches utilizing the same number of tokens for inference. The result demonstrates the importance of example absorption in internalizing essential information during the fine-tuning progress. In addition, when comparing \our with \our w/o Template Compression, we note that adding the template compression saves an average of 280 tokens across the datasets but experiences an average performance drop of 5\%. The result above demonstrates that while internalizing the template into model parameters reduces token usage, it requires a trade-off in terms of inference performance. 

\subsection{Analysis on Schedule Pattern}
In Table \ref{tab:ablation}, we test the effectiveness of different scheduling patterns during the progressive fine-tuning process, specifically focusing on how the decreasing speed curve influences the compression of the template and absorption of few-shot examples. The patterns tested include exponential, inverse-exponential, and linear decrease.

From the data in the table, we observe that the linear decreasing schedule delivers the most consistent and highest performance across all three evaluation metrics, indicating superior performance in both parsing efficiency and language model understanding. Conversely, the inverse-exponential schedule shows the least effectiveness, with scores considerably lower in all metrics compared to the linear schedule. The exponential decrease performs moderately, but still falls short of the linear schedule, suggesting that a steady, predictable reduction is more beneficial than more aggressive decrease. This analysis suggests that for adopting a linearly decreasing schedule for progressive fine-tuning may lead to better performance in terms of accuracy compared to other scheduling patterns.

\subsection{Analysis on Examples Retrieval Bank}
Table \ref{tab:ablation} examines the impact of varying proportion of the training set used for constructing relevant examples in the examples retrieval bank. The options tested include using 25\%, 50\%, and 100\% of the training set. The results show a trend where increasing the percentage of the training set used in the examples retrieval bank correlates with improved performance. This suggests that larger examples retrieval bank provides a richer set of few-shots for the model to learn from, thereby enhancing its ability to perform accurately across tasks.

\paragraph{Additional Experiments.} In Appendix \ref{appendix: progressive}, we explain why performing direct inference with a compressed prompt, while using the full prompt during training, is less effective compared to the progressive fine-tuning strategy employed by \our. Also, to provide a comprehensive analysis of \our's efficiency, we evaluate the inference speed and calculate the overall monetary cost during inference (details in Appendix \ref{appendix:speed}). In addition, to show that the principle of \our can generalize beyond the NL2Code domain, we evaluate \our on an additional task, which is presented in Appendix \ref{appendix: gsm8k}.

%% file: source/tables/direct_ft_res.tex

\begin{table*}[!t]
  \centering
  \caption{Comparison with direct fine-tuning baselines on NL2Code benchmark.}
  \label{tab:direct_ft res}
  \resizebox{\textwidth}{!}{
    \begin{tabular}{lcccccccccc}
    \toprule
    \rowcolor{gray!30} \multicolumn{11}{c}{\textbf{\textit{MBPP}}
    }  \\ 
    \midrule
    \multirow{2}{*}{\textbf{Model}} & \multicolumn{2}{c}{Template \textit{with} 5-shots} & \multicolumn{2}{c}{Template x0.6 \textit{with} 2-shots} & \multicolumn{2}{c}{Template Only}  & \multicolumn{2}{c}{No Template}& \multicolumn{2}{c}{\our} \\
    \cmidrule(lr){2-3} \cmidrule(lr){4-5} \cmidrule(lr){6-7} \cmidrule(lr){8-9} \cmidrule(lr){10-11}
    & Pass@1 & Input Tokens & Pass@1 & Input Tokens  & Pass@1 & Input Tokens  & Pass@1 & Input Tokens & Pass@1 & Input Tokens \\
    \midrule
        GPT-4 & 91.6 & 1181 & 87.4 & 424 & 87.3 & 226 & 77.2 & 43 & 87.9 & 43 \\ 
        GPT-3.5 & 82.7 & 1181 & 76.2 & 424 & 75.3 & 226 & 65.8 & 43 & 76.6 & 43 \\ 
        Mixtral-8x7B & 69.8 & 1263 & 65.8 & 453 & 65.7 & 238 & 56.3 & 54 & 66.3 & 54 \\ 
        Llama2-13B & 39.2 & 1286 & 37.5 & 471 & 36.4 & 251 & 26.4 & 58 & 37.1 & 58 \\ 
        Llama2-7B & 30.4 & 1286 & 27.7 & 471 & 27.3 & 251 & 18.3 & 58 & 27.9 & 58 \\ 
    
    \midrule
    \rowcolor{gray!30} \multicolumn{11}{c}{\textbf{\textit{NL2F}}
    }  \\ 
    \midrule
    \multirow{2}{*}{\textbf{Model}} & \multicolumn{2}{c}{Template \textit{with} 10-shots} & \multicolumn{2}{c}{Template x0.6 \textit{with} 5-shots} & \multicolumn{2}{c}{Template Only}  & \multicolumn{2}{c}{No Template}& \multicolumn{2}{c}{\our} \\
    \cmidrule(lr){2-3} \cmidrule(lr){4-5} \cmidrule(lr){6-7} \cmidrule(lr){8-9} \cmidrule(lr){10-11}
    & E.M. & Input Tokens & E.M. & Input Tokens  & E.M. & Input Tokens  & E.M. & Input Tokens &E.M. & Input Tokens \\
    \midrule
        GPT-4 & 94.8 & 3540 & 92.1 & 1838 & 89.7 & 680 & 82.5 & 286 & 91.6 & 286 \\ 
        GPT-3.5 & 85.5 & 3540 & 78.1 & 1838 & 76.2 & 680 & 70.4 & 286 & 78.4 & 286 \\ 
        Mixtral-8x7B & 69.3 & 4204 & 66.3 & 2191 & 63.8 & 814 & 54.2 & 339 & 65.2 & 339 \\ 
        Llama2-13B & 59.2 & 4202 & 54.9 & 2183 & 54.1 & 812 & 32.9 & 339 & 55.3 & 339 \\ 
        Llama2-7B & 45.4 & 4202 & 40.7 & 2183 & 38.5 & 812 & 21.8 & 339 & 40.8 & 339 \\ 

    \midrule
    \rowcolor{gray!30} \multicolumn{11}{c}{\textbf{\textit{NL2Bash}}
    }  \\ 
    \midrule
    \multirow{2}{*}{\textbf{Model}} & \multicolumn{2}{c}{Template \textit{with} 10-shots} & \multicolumn{2}{c}{Template x0.6 \textit{with} 5-shots} & \multicolumn{2}{c}{Template Only}  & \multicolumn{2}{c}{No Template}& \multicolumn{2}{c}{\our} \\
    \cmidrule(lr){2-3} \cmidrule(lr){4-5} \cmidrule(lr){6-7} \cmidrule(lr){8-9} \cmidrule(lr){10-11}
    & BLEU & Input Tokens & BLEU & Input Tokens  & BLEU & Input Tokens  & BLEU & Input Tokens & BLEU & Input Tokens \\
    \midrule
        GPT-4 & 86.7 & 1063 & 81.3 & 810 & 78.6 & 484 & 71.2 & 52 & 82.5 & 52 \\ 
        GPT-3.5 & 74.2 & 1063 & 67.5 & 810 & 65.1 & 484 & 61.2 & 52 & 67.7 & 52 \\ 
        Mixtral-8x7B & 63.8 & 1320 & 58.3 & 1053 & 54.9 & 603 & 47.6 & 68 & 57.2 & 68 \\ 
        Llama2-13B & 47.1 & 1244 & 43.9 & 988 & 41.6 & 574 & 35.1 & 64 & 43.5 & 64 \\ 
        Llama2-7B & 35.8 & 1244 & 32.7 & 988 & 31.4 & 574 & 22.1 & 64 & 31.6 & 64 \\ 

    \bottomrule
    \end{tabular}
  }
\end{table*}

%% file: source/tables/ablation_study/variants.tex
\begin{table*}[htbp]
  \centering
  \caption{Ablation Study of \our.}
  \label{tab:ablation_study_}
  \resizebox{\textwidth}{!}{
    \begin{tabular}{lccccccccc}
    \toprule
    \textbf{Methods} & \multicolumn{3}{c}{MBPP} & \multicolumn{3}{c}{NL2F} & \multicolumn{3}{c}{NL2Bash} \\
    \cmidrule(lr){2-4} \cmidrule(lr){5-7} \cmidrule(lr){8-10}
    (\textit{Inference on GPT-3.5}) & Pass@1 & Tokens & $1/\tau_{all}$ & E.M. & Tokens & $1/\tau_{all}$ & BLEU & Tokens & $1/\tau_{all}$ \\
    \midrule
    \textbf{\our} & 76.6 & \textbf{43} & 5.3x & 78.4 & \textbf{286} & 2.4x & 67.7 & \textbf{52} & 9.3x\\
    \arrayrulecolor{lightgray}\midrule
     \quad w/ $\tau_{tmp}$ = 0.3 & 78.1 & 107 & 2.1x & 81.4 & 407 & 1.7x & 70.5 & 241 & 2.0x\\
    \quad w/o Example Absorption & 72.9 & \textbf{43} & 5.3x & 73.5 & \textbf{286} & 2.4x & 64.6 & \textbf{52} & 9.3x \\
    \quad w/o Template Compression & \textbf{80.2} & 226 & 1.0x & \textbf{83.6} & 680 & 1.0x & \textbf{73.5} & 484 & 1.0x \\
    \arrayrulecolor{black}
    \bottomrule
    \end{tabular}
    }
\end{table*}

%% file: source/tables/ablation_study/schedule.tex
\begin{table}[htbp]
    \centering
    \caption{{Comparison of schedule pattern and example retrival bank of \our . The results are inferenced on GPT-3.5.}}
    \label{tab:ablation}
    \resizebox{0.5 \textwidth}{!}{
    \begin{tabular}{lccc}
    \toprule
    \textbf{\our} & MBPP(Pass@1) & NL2F(E.M.) & NL2Bash(BLEU) \\ 
    \midrule
    Pattern of Schedule $\mathcal{S}$   \\ 
    \quad - $\exp$ &  74.8 & 72.5 & 59.4 \\ 
    \quad - $\exp^{-1}$ & 67.3 & 64.9 & 52.8\\
    \quad - linear (\textbf{ours})  &  \textbf{77.6} & \textbf{78.4} & \textbf{67.7}\\ 
    \midrule
    Example Retrival Bank & \\ 
    \quad - 25\% & 75.9 & 77.5 & 66.2\\
    \quad - 50\% &  76.1 & 78.1 & 66.8\\ 
    \quad - 100\% (\textbf{ours}) & \textbf{77.6} & \textbf{78.4} & \textbf{67.7} \\  
    \bottomrule
    \end{tabular}
    }
\end{table}

%% file: source/6_discussion.tex
\section{Discussion}
\input{source/tables/time_overhead}
\subsection{Training Overhead of \our}
As \our is specifically designed to enhance the efficiency of LLMs during inference, the multi-stage design of progressive fine-tuning may introduce additional computational overhead. It is crucial to manage this overhead during the training phase to ensure that \our remains scalable and applicable in real-world, large-scale scenarios.


To illustrate the overhead of \our, we provide a detailed breakdown of the time overhead incurred during the training phase. We compare our method to the baseline \textit{template with k-shots}, which represents common use cases. Our evaluation includes the complete end-to-end process, covering dataset import and training times. All experiments were conducted on an A100x1-80G GPU. As shown in Table~\ref{tab: overhead}, \our consistently requires less time for both data preparation and training compared to the baseline. This efficiency is primarily due to the reduced number of input tokens per training instance during progressive fine-tuning. Furthermore, as outlined in our experimental settings, the number of training epochs for \our matches that of the direct fine-tuning baselines, ensuring no additional computational cost from extra training steps.

\subsection{General model ability}
Although progressive fine-tuning ensures high accuracy for domain-specific tasks, concerns may arise regarding whether this multi-step fine-tuning approach could negatively impact the overall model's generalizability. 
Several studies have examined fine-tuning for LLMs. \cite{wei2021finetuned} shows that multi-task fine-tuning enhances zero-shot and ICL capabilities. \cite{mosbach2023few} finds that few-shot fine-tuning preserves out-of-domain generalization similar to ICL settings. However, \cite{wang2022two} reveals that fine-tuning may overly adapt models to task-specific formats, reducing flexibility for new tasks. \cite{luo2023empirical} also explore catastrophic forgetting during continual fine-tuning.

In this work, \our is designed to enhance the efficiency of fine-tuned LLMs during inference while maintaining task-specific performance. Our approach is particularly suited for scenarios where LLMs are fine-tuned and deployed for domain-specific tasks. Although \our is currently limited to fine-tuning a single task, recent inference optimization techniques~\cite{ye2023aspen, wang2023multilora, sheng2023s} enable concurrent fine-tuning of multiple LoRA adapters while sharing a single backbone model. This allows for efficient adaptation to multiple tasks, reducing the number of parameters requiring fine-tuning. We will leave the integration of \our with these approaches in our future work.

We also provide a detailed discussion on preventing model overfitting in Appendix~\ref{appendix: over-fitting}.



%% file: source/tables/time_overhead.tex
\begin{table*}[htbp]
    \centering
    \caption{Time Overhead of \our during fine-tuning.}
    \label{tab: overhead}
    \resizebox{\textwidth}{!}{
    \begin{tabular}{lcccccc}
        \toprule
        \multirow{2}{*}{\textbf{Model}} & \multicolumn{2}{c}{\textbf{MBPP}} & \multicolumn{2}{c}{\textbf{NL2F}} & \multicolumn{2}{c}{\textbf{NL2Bash}} \\ 
        \cmidrule(lr){2-3} \cmidrule(lr){4-5} \cmidrule(lr){6-7}
         & \our & Template \textit{with} 5-shots & \our & Template \textit{with} 10-shots & \our & Template \textit{with} 10-shots \\ \midrule
        GPT-4 & 01h 38m 54s & 02h 13m 07s & 03h 46m 15s & 04h 32m 12s & 03h 23m 18s & 04h 17m 28s \\ 
        GPT-3.5 & 01h 19m 03s & 01h 48m 55s & 03h 02m 29s & 03h 44m 46s & 02h 23m 19s & 03h 09m 16s \\ 
        Llama2-13B & 00h 45m 27s & 01h 14m 10s & 01h 36m 17s & 02h 16m 21s & 01h 12m 48s & 01h 48m 03s \\ 
        \bottomrule
    \end{tabular}
    }
\end{table*}

%% file: source/5_conclusion.tex
\section{Conclusion}
In this paper, we propose \our, a prompt internalization method that internalizes repetitive prompt knowledge into LLMs parameters. We develop specific compression strategies for different components of the prompt, accompanied by a tailored progressive fine-tuning pipeline. Extensive experiments demonstrate that our method maintains comparable performance effectiveness while accelerating inference speed with less token usage.

\section{Limitations}
Through extensive experiments in this paper, \our has demonstrated a strong ability to reduce model costs during inference. However, as shown in Table~\ref{tab:direct_ft res}, \our still exhibits a performance gap in accuracy compared to the use of original prompts (i.e., \textit{template with k-shots}).
Also, while we empirically validate the effectiveness of \our, a theoretical analysis of model parameter updates and the training pipeline complexity is still required. 
In addition, although the principle of \our can be generalized to most downstream NLP tasks (as demonstrated in Appendix \ref{appendix: gsm8k}, further empirical verification is needed on more advanced tasks. In future work, we plan to conduct more evaluations on several complex tasks, including long-document summarization, question-answering in specialized technical domains, etc.


\section{Ethics Statement}
All datasets used in this paper are publicly available and have been reviewed to ensure they do not contain any personally identifiable information or offensive content. Additionally, experiments were conducted on computational clusters with NVIDIA A100 GPUs. It is important to note that this could have an environmental impact, and the carbon footprints were monitored in real time.

%% file: source/appendix.tex
\section{Additional Experiments} \label{appendix:add_exp}
\subsection{Dataset Details} \label{appendix:dataset}

\paragraph{MBPP} The MBPP dataset, as detailed by \cite{mosbach2023few}, consists of Python programming tasks, each accompanied by a description in natural language that has been expertly curated. The dataset is segmented into training and test sets, with 974 and 102 examples, respectively.
\paragraph{NL2F} The NL2F dataset, as detailed by \cite{zhao2024nl2formula}, consists of 70,799 pairs of NL queries and spreadsheet formulas and covers 21,670 tables. We follow the dataset instructions \cite{zhao2024nl2formula} to randomly split data into a training set (75\%), validation set (10\%), and test set (15\%).
\paragraph{NL2Bash} The NL2Bash dataset, as described by \cite{lin2018nl2bash}, comprises snippets of Bash code, each paired with a natural language description expertly curated. The dataset is divided into training and test sets, containing 8,090 and 606 examples, respectively.

\subsection{Implementation Details} 
\label{appendix:impl_detail}
\paragraph{Fine-tuning Procedures} For \our training, we adopt LoRA \cite{hu2021lora} with a rank of 32. For GPT-series and open-source model fine-tuning we train models for MBPP/NL2F/NL2Bash with 6/12/12 epochs, 16/128/128 batch size, 200/200/200 checkpoint interval, and 4096/4096/4096 context window length, respectively.
\paragraph{Model Inference} We provide the detailed parameters we adopted during fine-tuned LLM inference: temperature equal to 0, max tokens equal to 1028, top p equal to 0.95, presence penalty equal to 0, and frequency penalty equal to 0.
\input{source/tables/appendix/gist}
\paragraph{Baseline Settings} For prompt compression baselines comparison, we set the template compression ratio $\tau_{tmp} = 0.3$. For direct fine-tuning baselines, we apply LLMLingua-2 \cite{pan2024llmlingua} as the default template compressor as it performs the best in Table \ref{tab:baseline_compare}.

\noindent 
\\
\textbf{Parameter Settings for \our} 
\\
\noindent
1) Number of top-k for example absorption: We set the initial k as 5/10/10 across MBPP/NL2F/NL2Bash for the initial number of few-shot examples for example absorption. During progressive fine-tuning, we decrease k linearly in the order of 5-2-0/10-5-0/10-5-0 across MBPP/NL2F/NL2Bash.
\\
\noindent
2) Number of $\tau_{tmp}$ for template compression: For the prompt compression baseline experiments, we set the final template rate to 0.3, which is used in the last stage of fine-tuning as well as inference. For the other experiments and ablation studies, we set the final template rate to 0 to achieve full internalization.
\input{source/tables/appendix/speed}
\paragraph{Cost Evaluation} We compute the total costs based on the price shown in OpenAI Pricing\footnote{https://openai.com/api/pricing/}

\paragraph{Computational Resource} We conduct all experiments on AzureAI Machine Learning Studio with one A100x1-80G computational cluster

\subsection{Comparison with Gist Tokens} \label{appendix:gist}
We report the comparison result of \our with Gist Tokens \cite{mu2024learning} on Table \ref{tab:gist}. 
Gist Tokens showcases consistent performance, with notable results in NL2Bash where it achieves a BLEU score of 22.7, suggesting a moderate alignment with the dataset’s requirements. In contrast, \our demonstrates superior performance across all metrics and datasets, particularly excelling in the NL2Bash dataset with a BLEU score of 31.6 and maintaining similar efficiency in token usage. The results demonstrate that our approach significantly outperforms the Gist token while conducting overall the same compression rate.

\subsection{Effectiveness of progressive fine-tuning in \our}
\label{appendix: progressive}
In this experiment, we compare \our with the method of direct fine-tuning with a full-loaded prompt (template plus few-shot examples) followed by inferencing with queries only (designated as \textit{Template with k-shots*} in table \ref{tab: progressive}). We use the comparison to demonstrate the effectiveness of progressive fine-tuning for updating model parameters properly. The result is shown in Table \ref{tab: progressive}. We can clearly observe that 
\input{source/tables/appendix/effect_progressive_ft} \textit{Template with k-shots*} has a large performance degradation compared to \our. This indicates that fine-tuned LLMs struggle to establish a proper connection between the full-length prompts used in training and the query-only prompts used during inference. It also motivates the development of the progressive fine-tuning strategy in \our. Beyond empirical experiments demonstration, we will leave the theoretical proof of the effectiveness of \our (Algorithm \ref{alg:inprompt_training}) in our future work.

\begin{figure*}[htbp]
    \centering
    \includegraphics[width=1\linewidth]{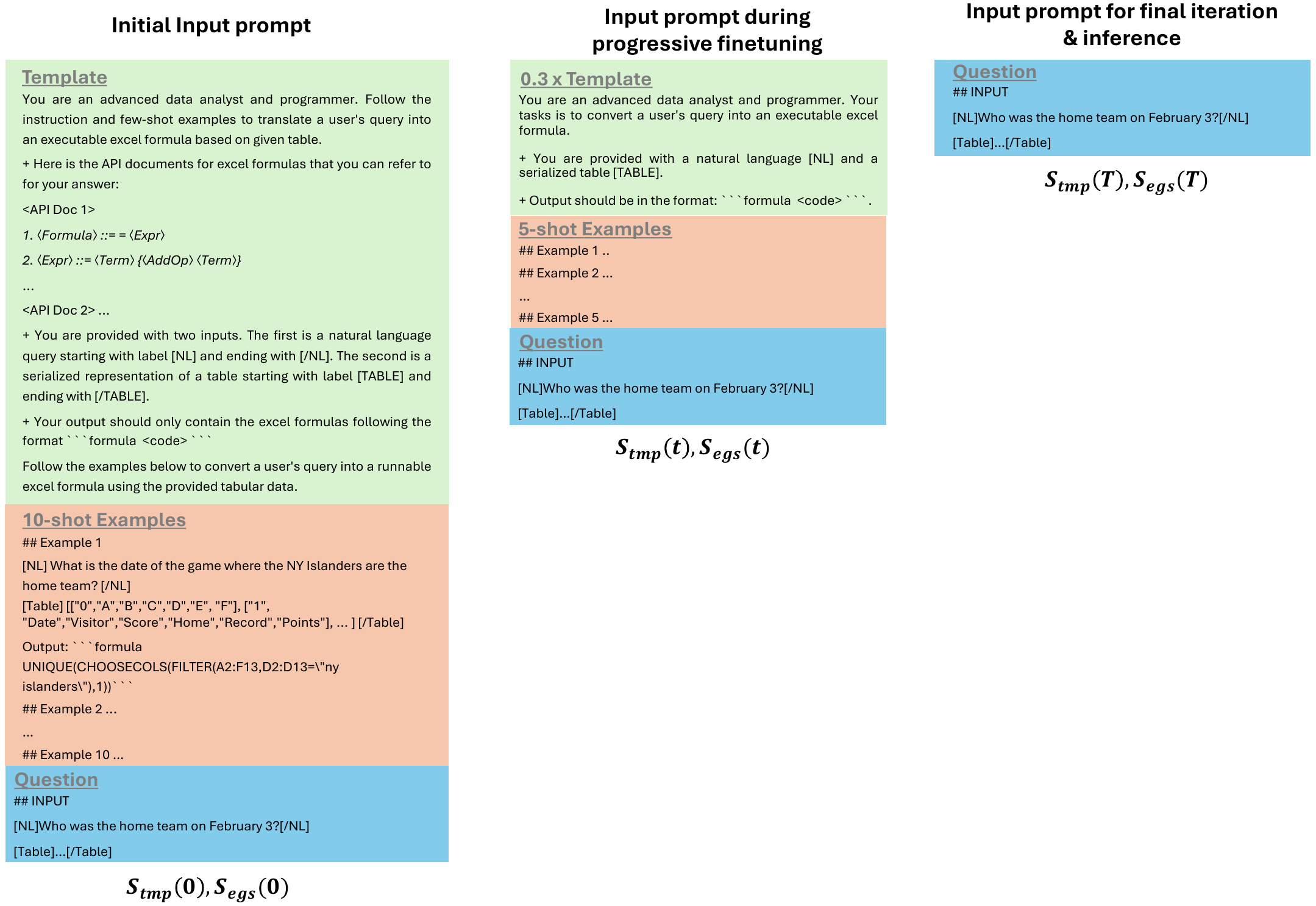}
    \caption{{ An Example from NL2F demonstrating how an original prompt is preprocessed through template compression and example absorption in \our for progressive fine-tuning and final inference.
    }}
    \label{fig:preprocess_example}
\end{figure*}
\subsection{Experiments on Inference Speed} \label{appendix:speed}
The experimental results presented in Table \ref{tab:speed} illustrate the low latency characteristics of \our during inference across three datasets, MBPP, NL2F, and NL2Bash. Specifically, for the MBPP dataset, \our achieves an inference speed of 4.17 s/instance on the GPT-4 model, closely aligning with the 4.36 s/instance observed in the no template setup and far surpassing the more resource-intensive template with 5-shots configuration at 10.21 s/instance. In the NL2F dataset, \our similarly demonstrates its efficiency with an inference speed of 2.15 s/instance for GPT-4, which is nearly equivalent to the 2.12 s/instance observed without any template and significantly outperforms the elaborate template with 10-shots configuration, which achieves 12.47 s/instance. The experimental results outlined in the table also highlight the efficiency of \our in the NL2Bash dataset. Notably, for GPT-4 under the NL2Bash benchmark, \our maintains a competitive inference speed of 1.18 s/instance, matching the performance seen in the no template scenario and markedly better than the template with 5-shots setup, which records a slower speed of 4.46 s/instance. The result across three NL2Code benchmarks highlights \our’s capability to maintain competitive inference speeds while minimizing latency efficiently. 

\subsection{Evaluating \our on GSM8K}
\label{appendix: gsm8k}
To demonstrate the generalization ability of \our on other domain tasks, we also test on the GSM8K dataset \cite{cobbe2021training}. We use the same experiment settings stated in our experiment setups to compare with the prompt compression baselines. The results are shown in Table \ref{tab: gsm8k}. The result demonstrates that, under the same compression rate for inference, \our outperforms other compression baselines by 3\%-34\% for the arithmetic reasoning task.

\input{source/tables/appendix/gsm8k}
\section{Additional Discussions}
\subsection{Prevention of Model Over-fitting}
\label{appendix: over-fitting}
To prevent LLMs from over-fitting due to lengthy input prompts, such as long templates, and to mitigate over-fitting during multi-stage fine-tuning, we have implemented several strategies:
\begin{itemize}[leftmargin=*,itemsep=0pt]
    \item \textbf{Prompt Selection:} For each dataset, we utilize default prompts provided by the authors or sourced from widely recognized papers. In our experiments, these prompts are applied for the baseline \textit{template with 5/10 shots}, which outperforms other approaches and aligns with results from sources like Papers With Code. This ensures that our prompts will not cause performance degradation due to over-fitting input.
    
    \item \textbf{Model Checkpointing:} To mitigate over-fitting during fine-tuning, particularly due to excessively lengthy training epochs, we implemented a model checkpointing strategy. We save the model state every 200 training steps and evaluate each checkpoint on a separate validation dataset. This allows us to track performance changes throughout the whole training process. By comparing checkpoints, we identify the optimal iteration that achieves the best results, determining the appropriate number of training steps and epochs for our experiments.

    \item \textbf{Validation Monitoring:} Training is halted using an early stopping technique when the validation loss begins to rise or ceases to decline, indicating potential over-fitting. Additionally, we manually monitor the training and validation losses against the number of training steps for each experiment of \our. This visualization helps ensure that each model avoids training over-fitting by providing a clear depiction of the training dynamics and enabling timely adjustments.
\end{itemize}

\section{Example Demonstration} \label{appendix: example_demo}
We demonstrate an example of how we schedule and pre-process an input prompt through both template compression and example absorption in Figure \ref{fig:preprocess_example}. 
During the initial fine-tuning phase, the input prompt will fully incorporate the template and 10-shot examples for the NL2F dataset. After a specified number of training iterations, the template will undergo compression at a rate of 0.3, and the number of examples will be reduced to five. This modified prompt is then used for the intermediate stage of fine-tuning. In the final phase, the template and few-shot examples are removed from the training prompt. It is important to note that the query remains unchanged throughout the entire progressive fine-tuning process. The prompt used in the last stage, which consists solely of the query, will also serve as the input for subsequent model inference. This method enables the fine-tuned language model to perform zero-shot inference without the need for an instruction or document template.
\section{Prompts}
\subsection{GPT-4 Generation (Baseline) instruction} \label{appendix:gpt}
\begin{figure}[htbp]
    \centering
    \includegraphics[width=1\linewidth]{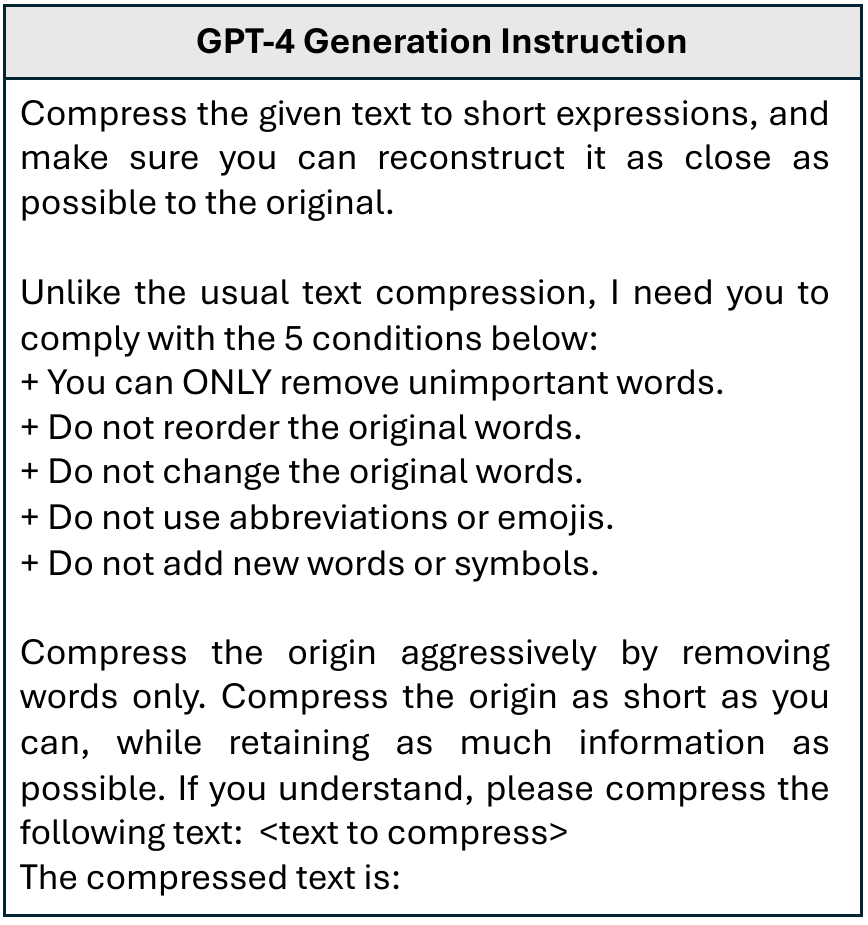}
    \caption{{ The instruction baseline for the baseline method GPT-4 Generation. 
    }}
    \label{fig:gpt-4_generation}
\end{figure}

For detailed prompts, please refer to Figure \ref{fig:gpt-4_generation}.
\subsection{Prompts of \our on NL2Code}
\begin{figure*}[!t]
    \centering
    \includegraphics[width=1\linewidth]{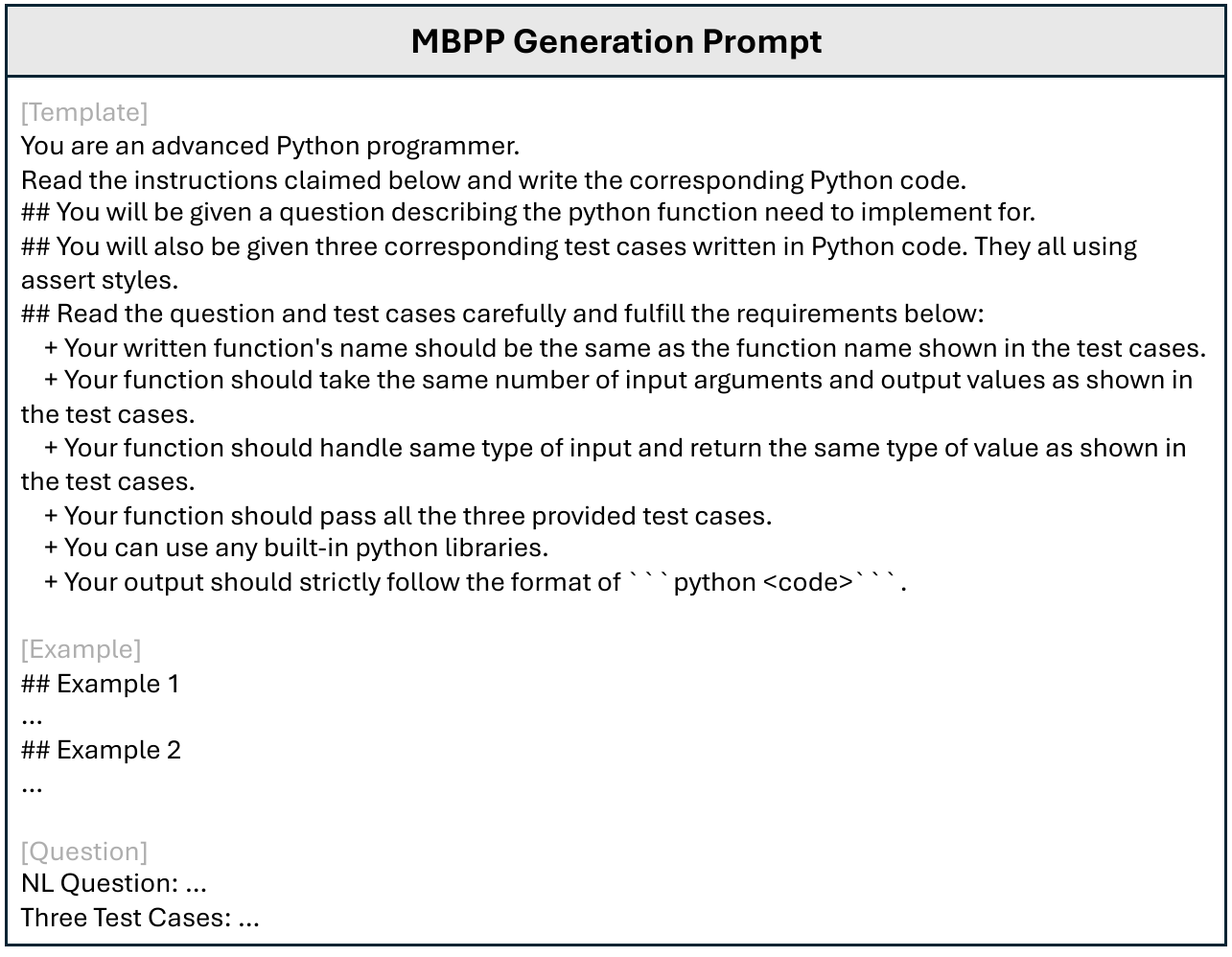}
    \caption{{ Prompts of MBPP
    }}
    \label{fig:prompt_mbpp}
\end{figure*}

\begin{figure*}[!t]
    \centering
    \includegraphics[width=1\linewidth]{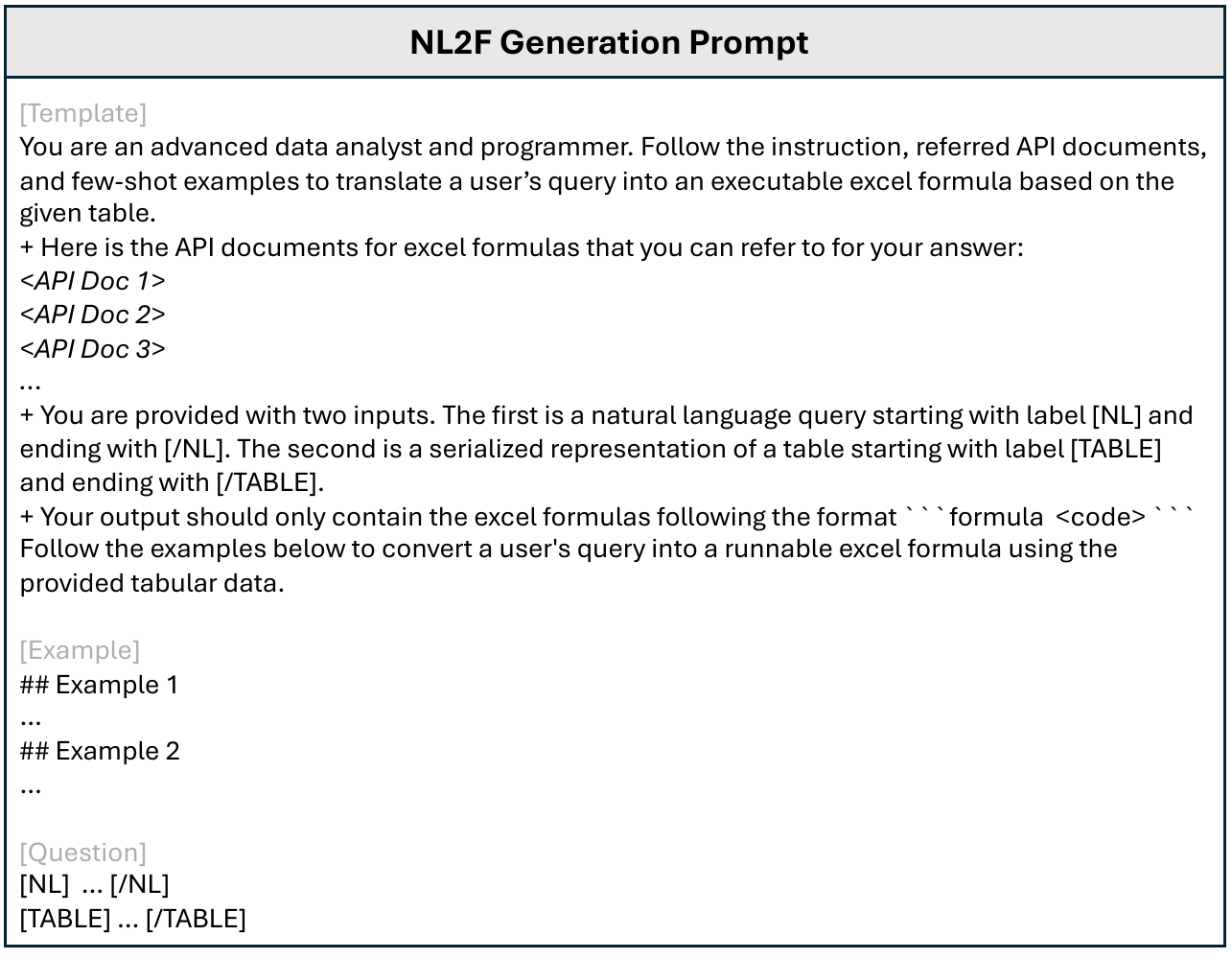}
    \caption{{ Prompts of NL2F
    }}
    \label{fig:prompt_nl2f}
\end{figure*}

\begin{figure*}[!t]
    \centering
    \includegraphics[width=1\linewidth]{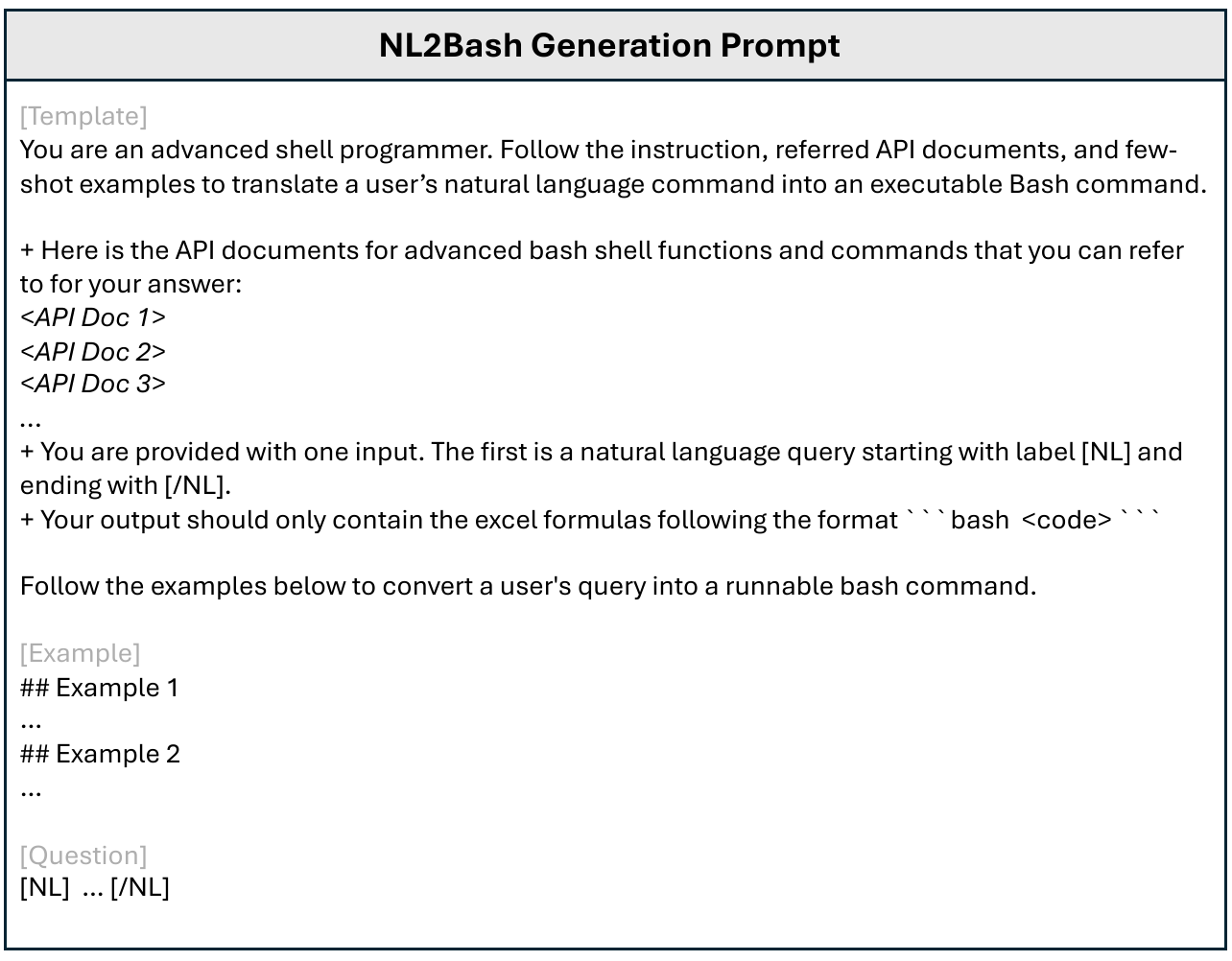}
    \caption{{ Prompts of NL2Bash
    }}
    \label{fig:prompt_nl2bash}
\end{figure*}

For detailed prompts of each dataset, please refer to Figures \ref{fig:prompt_mbpp},\ref{fig:prompt_nl2f},\ref{fig:prompt_nl2bash}. 
\label{sec:appendix}

%% file: source/tables/appendix/gist.tex
\begin{table*}[!t]
  \centering
  \caption{Comparison with Gist Tokens \cite{mu2024learning}}
  \label{tab:gist}
  \resizebox{\textwidth}{!}{
    \begin{tabular}{lccccccccc}
    \toprule
    \textbf{Methods} & \multicolumn{3}{c}{MBPP} & \multicolumn{3}{c}{NL2F} & \multicolumn{3}{c}{NL2Bash} \\
    \cmidrule(lr){2-4} \cmidrule(lr){5-7} \cmidrule(lr){8-10}
    (\textit{Inference on Llama2-7B}) & Pass@1 & Tokens & $1/\tau_{all}$ & E.M. & Tokens & $1/\tau_{all}$ & BLEU & Tokens & $1/\tau_{all}$ \\
    \midrule
    Gist Tokens & 10.2 & 61 & 4.1x & 17.5  & 342 & 2.4x & 22.7 & 66 & 8.6x \\
    \midrule
    \our & 27.9 & 58 & 4.3x & 40.8 & 339 & 2.4x & 31.6 & 64 & 9.0x\\
    \bottomrule
    \end{tabular}
  }
\end{table*}

%% file: source/tables/appendix/speed.tex
\begin{table*}[!ht]
    \centering
    \caption{Speed (s/instance) Comparison of \our with Direct Fine-tuning baseline on NL2Code benchmarks.}
    \label{tab:speed}
    \resizebox{\textwidth}{!}{
    \begin{tabular}{lccccc}
    \toprule
    \textbf{Model} & Template \textit{with} 5-shots & Template x0.6 \textit{with} 2-shots & Template  & No template & \our \\ \midrule
    \rowcolor{gray!30} \multicolumn{6}{c}{\textbf{\textit{MBPP}}
    }  \\ 
    \midrule
    GPT-4 & 10.21 & 8.68 & 7.29 & 4.36 & \textbf{4.17} \\ 
    GPT-3.5 & 5.43 & 3.68 & 3.06 & 1.35 & \textbf{1.31} \\ 
    Mixtral-8x7B & 4.84 & 3.23 & 3.14 & 1.76 & \textbf{1.62} \\ 
    Llama2-13B & 3.17 & 2.54 & 2.19 & 1.08 & \textbf{1.13} \\ 
    Llama2-7B & 2.95 & 2.27 & 1.95 & 0.84 & \textbf{0.76} \\ 

    \midrule
    \rowcolor{gray!30} \multicolumn{6}{c}{\textbf{\textit{NL2F}}
    }  \\ 
    \midrule
    GPT-4 & 12.47 & 8.43 & 4.16 & 2.12 & \textbf{2.15} \\ 
    GPT-3.5 & 8.16 & 5.26 & 2.18 & 1.46 & \textbf{1.44} \\ 
    Mixtral-8x7B & 6.27 & 4.71 & 3.17 & 1.19 & \textbf{1.20} \\ 
    Llama2-13B & 4.15 & 2.95 & 1.25 & 0.63 & \textbf{0.63} \\ 
    Llama2-7B & 3.83 & 2.03 & 1.24 & 0.41 & \textbf{0.39} \\ 

    \midrule
    \rowcolor{gray!30} \multicolumn{6}{c}{\textbf{\textit{NL2Bash}}
    }  \\ 
    \midrule
    GPT-4 & 4.46 & 3.18 & 1.57 & 1.18 & \textbf{1.18} \\ 
    GPT-3.5 & 2.79 & 2.36 & 1.29 & 1.02 & \textbf{1.02} \\ 
    Mixtral-8x7B & 2.65 & 2.23 & 1.59 & 1.13 & \textbf{1.13} \\ 
    Llama2-13B & 2.21 & 1.96 & 1.41 & 1.05 & \textbf{1.05} \\ 
    Llama2-7B & 1.86 & 1.49 & 1.02 & 0.87 & \textbf{0.87} \\ 
    \bottomrule
    \end{tabular}
    }
\end{table*}




%% file: source/tables/appendix/effect_progressive_ft.tex
\begin{table}[!ht]
    \centering
    \caption{Demonstration of progressive fine-tuning in \our}
    \label{tab: progressive}
    \resizebox{\linewidth}{!}{
    \begin{tabular}{lccc}
        \toprule
        \textbf{Dataset (Inference on GPT3.5)} & \textbf{MBPP} & \textbf{NL2F} & \textbf{NL2Bash} \\ \midrule 
        Template with k-shots$^*$ & 69.1 & 71.7 & 63.2 \\ 
        PromptIntern & \textbf{76.6} & \textbf{78.4} & \textbf{67.7} \\ 
        \bottomrule
    \end{tabular}
    }
\end{table}

%% file: source/tables/appendix/gsm8k.tex
\begin{table}[!ht]
    \centering
    \caption{Comparison of prompt compression baselines on GSM8K.}
    \label{tab: gsm8k}
    \resizebox{\linewidth}{!}{
    \begin{tabular}{lcc}
        \toprule
        \textbf{Methods (Inference on GPT 3.5)} & \textbf{E.M.} & \textbf{Tokens} \\ \midrule
        Selective Context & 63.5 & 443 \\ 
        LLMLingua & 83.2 & 452 \\ 
        LLMLingua-2 & 81.9 & 458 \\ 
        \our & \textbf{85.7} & 458 \\ 
        \bottomrule
    \end{tabular}
    }
\end{table}